\documentclass[11pt]{article}

\usepackage{hyperref}
\usepackage{latexsym}
\usepackage{graphicx}
\usepackage{epsfig}
\usepackage{xurl}
\usepackage{amsmath,amssymb}
\numberwithin{equation}{section}
\usepackage{tabularx}
\usepackage{authblk}
\usepackage{mathrsfs}
\usepackage{algorithm}
\usepackage{soul}
\usepackage{multirow}
\usepackage{color}
\usepackage{subcaption}
\usepackage{enumerate}
\usepackage{booktabs}
\usepackage{svg}
\usepackage{cleveref}
\usepackage{xcolor}
\usepackage{capt-of}
\usepackage{makecell}
\usepackage{cite}
\usepackage[table]{xcolor}
\usepackage[noend]{algpseudocode}
\usepackage[utf8]{inputenc}

\newcommand{\name}{GTCL}

\title{Latent Trajectory Discrimination for AI-Generated Text Detection}

\author{
Gianluca Bonifazi$^{1}$,
Christopher Buratti$^{1}$,
Michele Marchetti$^{1}$,
Federica Parlapiano$^{1}$,
Giulia Quaglieri$^{1}$,
Davide Traini$^{1,2}$,
Domenico Ursino$^{1}$,
Luca Virgili$^{1,*}$ \\
$^{1}$ DII, Polytechnic University of Marche \\
$^{2}$ CHIMOMO, University of Modena and Reggio Emilia \\
$^{*}$ Corresponding author \\

g.bonifazi@univpm.it;
c.buratti@pm.univpm.it;
michele.marchetti@univpm.it;
g.quaglieri@pm.univpm.it;
f.parlapiano@pm.univpm.it;
davide.traini@unimore.it;
d.ursino@univpm.it;
luca.virgili@univpm.it

}

\date{}

\begin{document}
\maketitle

\begin{abstract}
Most existing approaches to AI-Generated Text Detection (AIGTD) treat documents as static objects and base their decisions on aggregate statistics or globally compressed embeddings. However, this perspective overlooks the inherently dynamic nature of autoregressive generation, where content evolves progressively through the latent space. In this paper, we reformulate AIGTD as the problem of distinguishing between latent generation trajectories. Instead of relying on static representations, we model how textual representations evolve across the sequence. To this end, we propose Geometric Trajectory and Contrastive Learning (\name{}), a framework that segments the document into ordered local units, encodes each unit in an embedding space, and constructs a structured and sequence-level representation. GTCL then applies contrastive learning to these trajectories to learn geometric regularities associated with the autoregressive generation. Evaluations performed on three different benchmarks and several approaches show that GTCL outperforms detection baselines consistently, which implies that explicitly modeling sequential dynamics provides robust discriminative signals across models and domains. These results suggest that modeling trajectory differences could improve detection and open up a dynamic direction that has been underexplored in previous AIGTD literature.
\end{abstract}


\section{Introduction}
\label{sec:Introduction}

The proliferation of Large Language Models (LLMs) has radically altered the production of digital content, making synthetic texts increasingly indistinguishable from human-generated ones. This development has made AI-Generated Text Detection (AIGTD) a central issue \cite{Yang*25}, with significant implications in academia, where the widespread use of generative tools raises new challenges regarding authorship and scientific integrity \cite{Pudasaini*24}.

Over time, the literature has identified three categories of defensive strategies \cite{Yang*24-2}, i.e., supervised methods, zero-shot approaches, and watermarking techniques. Supervised methods involve training classifiers on labeled corpora; they achieve high performance under controlled conditions but often degrade in cross-model scenarios. Zero-shot approaches exploit intrinsic probabilistic signals, such as perplexity or log-probability curvature, without requiring specific training. Finally, watermarking techniques identify signals directly into the generative process but require control of the model and are not applicable to arbitrary texts.

Despite their methodological differences, these paradigms share the implicit assumption that text is a static object. Decisions are based on aggregated statistics or compressed embeddings at the document level without explicitly modeling how latent representations evolve along the sequence. As the superficial differences between human and synthetic texts diminish, this static perspective reveals structural limitations. Empirical analyses have highlighted significant instability in the performance of existing detectors \cite{Bellini*24}, while theoretical results link maximum achievable accuracy to the Total Variation distance between the distributions of human and generated texts \cite{Sadasivan*25}. Furthermore, adversarial tests show that recursive paraphrasing strategies can significantly reduce the effectiveness of existing systems\cite{Cheng*25-2}. This evidence suggests that the distinctive signature of autoregressive generation lies more in the dynamic structure of the generative process than in isolated lexical artifacts. In other words, differences may emerge in how semantic representations progressively evolve as a text progresses.

In this paper, we test the aforementioned hypothesis by introducing the Geometric Trajectory and Contrastive Learning (\name) framework. Rather than treating a document as a static object summarized by a single embedding, GTCL represents it as a trajectory in latent space. To achieve this, it segments the text into fixed-length overlapping windows, encodes each window as a feature vector, and organizes the resulting feature vectors into a structured matrix that preserves sequential order. In this formulation, a document is modeled as a sequence of local semantic states, and the progression of these states defines a discrete latent dynamic.

GTCL is based on different key design principles. First, it shifts the unit of analysis from global pooled representations to trajectory-based modeling. This allows it to explicitly capture the evolution of semantic states along the sequence. Then, it operates on consecutive feature vector differences to model the first order discrete dynamics of the trajectory rather than its absolute positions. Next, it defines a temporally aligned similarity measure over these differences and applies supervised contrastive learning \cite{Khosla*20} directly at the trajectory level. This design allows GTCL to learn structural regularities associated with autoregressive generation without requiring access to logits, decoding traces, or watermarking signals.

The contributions of this paper can be summarized as follows:

\begin{itemize}

    \item It reformulates AIGTD as a discrimination problem over trajectory differences rather than on static document representations.

    \item It introduces a temporally aligned similarity measure based on first order embedding differences that captures directional and locally structured semantic transitions.

    \item It proposes \name{}, a trajectory-level, supervised, contrastive learning-based framework that structures the projection space according to generation-specific geometric dynamics.

    \item It provides empirical validation on adversarial and multi-model benchmarks, including the RAID dataset \cite{Dugan*24}, the NYT-AI dataset \cite{Roy*25}, and an OpenReview-based dataset. It shows that GTCL outperforms detection baselines across all evaluation settings.
    
\end{itemize}


The rest of this paper is organized as follows: Section~\ref{sec:Related-Work} provides a review of the existing literature. Section~\ref{sec:Technical-Description-GTCL} formally presents the \name{} framework. Section~\ref{sec:Experiments} describes the experimental setup and results. Section~\ref{sec:Discussion} discusses the implications of the proposed framework. Finally, Section~\ref{sec:Conclusion} summarizes our contributions and outlines future work.

\section{Related work}
\label{sec:Related-Work}

Research on AIGTD can be organized into three main paradigms, namely: {\em (i)} zero-shot or training-free approaches, {\em (ii)} supervised, training-based approaches, and {\em (iii)} watermarking-based approaches. In this section, we analyze these lines of research and emphasize that none of them explicitly model text as a sequential trajectory in latent space. In particular, we analyze a line of research per subsection.

\subsection{Zero-shot detection based on probability distributions}

Initial studies primarily fall within the zero-shot paradigm \cite{Mao*24,Shi*24}, which exploits the statistical properties of language model probability distributions to detect synthetic text \cite{SuLv25, LvSu26}. Classic indicators include perplexity, entropy, and mean log-likelihood; all these indicators capture the greater predictability of generated text.

DetectGPT \cite{Mitchell*23} makes a significant conceptual advance by observing that generated texts tend to lie in regions of negative curvature in the log-likelihood landscape. It measures probability variation after small perturbations using the generative model's intrinsic geometric properties. DetectGPT4Code \cite{Yang*23-2} extends this procedure to code, but it requires domain-specific adaptations. Other training-free approaches, such as Binoculars \cite{Hans*24}, compare the distributions of two models. DNA-GPT \cite{Yang*24}, on the other hand, uses n-gram divergences based on regenerating truncated segments. 

While these methods leverage the internal properties of language models, they primarily rely on aggregate statistics or local perturbations.

\subsection{Supervised methods}

As LLMs have become more complex, research has progressively shifted toward supervised approaches \cite{Wang*23-3,Wu*23-2}, in which a classifier is trained on labeled corpora of human- and AI-generated texts. Some studies have explored the direct application of LLMs as zero-shot classifiers \cite{BhLi24}. They have also highlighted structural limitations of this procedure, such as self-recognition bias and inconsistent evaluation results.

Most supervised methods rely on encoder-only transformer architectures, such as RoBERTa \cite{Guo*23} or DeBERTaV3 \cite{HeGaCh23}, which are fine-tuned with a sequence-level classification head. In some cases, these architectures are integrated with sequential or convolutional components \cite{Mo*24}. More recent models, such as ModernBERT \cite{Warner*25, DrYiLa25}, improve long context handling and computational efficiency. Despite these architectural developments, the paradigm remains static because the document is compressed into an aggregated representation (e.g., via CLS token pooling), on which binary classification is performed. The order and internal dynamics of latent representations are not explicitly modeled as a structural object.

Extensions based on contrastive learning \cite{Hu*24} and adversarial training have been proposed to address the problems of overfitting and poor cross-model or cross-domain generalization. RADAR \cite{HuChHo23} formulates detection as an adversarial game between the detector and the paraphraser to induce robustness against superficial rewrites. DeTeCtive \cite{Guo*24} uses multi-level contrastive learning to distinguish between model families, treating each LLM as a unique author. Approaches such as ConDA \cite{Bhattacharjee*23} aim for domain-invariant representations, while the approaches in \cite{LaCoTa24} apply contrastive objectives at the document or span level.

These strategies improve separability in latent space and robustness to adversarial transformations. However, the unit of representation remains static, consisting of document embeddings, paragraphs, or individual text portions. Even when considering local segments, the latter are not interpreted as part of an ordered trajectory in embedding space. The encoder model implicitly handles sequentiality, but does not explicitly model it as an evolutionary dynamic of the generative process.

\subsection{Watermarking}

Watermarking approaches \cite{Hao*25,Qu*25,Kirchenbauer*24, Li*26} adopt a conceptually different strategy. Rather than inferring the text's origin from statistical regularities, they introduce an identifying signal when text is generated. Recent techniques modulate the sampling distribution \cite{Kirchenbauer*23} to insert verifiable signatures while maintaining the text's fluidity.

However, watermarking requires control over the generative process, limiting its application to white-box scenarios. Additionally, the robustness of the watermark may decrease in the presence of paraphrasing or rewriting. Consequently, this paradigm does not address the problem of post-hoc detection on arbitrary texts.

\subsection{Beyond static representations}

Recent benchmarks, such as RAID \cite{Dugan*24}, have revealed that detectors are susceptible to paraphrasing and domain variations. RAID evaluations reveal that detectors struggle not because of individual lexical choices, but because the detectability of generated text comes from the global properties of the generative process. These properties include sampling strategies, repetition penalties, and adversarial modifications. Current methods fail to robustly capture them.

Recent analyses of the probabilistic dynamics of autoregressive generation confirm this insight as well. In \cite{Sun*26}, the authors show that traditional zero-shot methods, based on aggregate statistics over the entire sequence, neglect relevant temporal variations. This highlights the systematic differences between human- and AI-generated text in the final stages of the sequence. This result suggests that the evolutionary structure of text contains discriminative signals that global representations do not capture.

Despite this evidence, literature continues to treat text as predominantly static, analyzing it through aggregate statistics or compressed embeddings at the document level. To the best of our knowledge, there are no approaches that explicitly model documents as ordered trajectories of latent representations or apply contrastive learning directly to sequential dynamics in embedding spaces.

GTCL aims to address this methodological gap. Rather than comparing texts using static representations, it models each document as a geometric trajectory in a high-dimensional space constructed from ordered local windows. Then, it compares these trajectories using a sequence-level contrastive objective.

\section{The \name{} framework}
\label{sec:Technical-Description-GTCL}

This section provides a technical description of \name{}. Specifically, Subsection \ref{sub:Motivation} explains the motivation behind our framework. Subsection \ref{sub:Overview-GTCL-Behavior} provides an overview of its workflow. Finally, Subsection \ref{sub:Approach-Underlying-GTCL} details the various steps that \name{} takes to achieve its objectives.

\subsection{Motivation}
\label{sub:Motivation}

Modern LLMs generate text through an autoregressive mechanism, where each new token is conditioned on the previously generated context \cite{Vaswani*17}. This process introduces strong local dependencies and is expected to induce locally consistent transitions in the latent representation space \cite{Holtzman*20, Mitchell*23}. While human-written text also exhibits semantic coherence, it is not constrained by a single predictive objective and may naturally include revisions, shifts in perspective, stylistic variations, and semantic discontinuities \cite{GeStRu19}. As a result, the evolution of latent representations in human-written documents is expected to exhibit greater directional variability than that of AI-generated texts.

Most existing AI-generated text detection methods treat documents as static objects, making predictions from aggregate statistics or a single compressed document representation. Although effective in many settings, these approaches overlook how semantic content evolves throughout a document. To address this limitation, GTCL models a document as an ordered sequence of partially overlapping text segments, each mapped into a shared latent space. The resulting sequence of embeddings is interpreted as a discrete trajectory, whose local transitions describe the semantic evolution of the text. By differentiating it, it is also possible to focus on the local evolution of the text, instead of considering only its absolute schema.

In this formulation, detecting AI-generated text becomes a matter of distinguishing differences in trajectories rather than in static document representations. The underlying hypothesis is that differences between human- and AI-generated texts are reflected in the geometry of semantic transitions rather than in isolated lexical features. Accordingly, GTCL performs supervised contrastive learning directly on trajectory dynamics while constructing contrastive groups of hard positive and negative examples. This approach enables the learned representation to more effectively distinguish between semantically similar documents from different classes and reduce variability among challenging samples within a class. 

\subsection{Overview of the \name{} workflow}
\label{sub:Overview-GTCL-Behavior}

\begin{figure}[ht!]
    \centering
    \includegraphics[width=\textwidth]{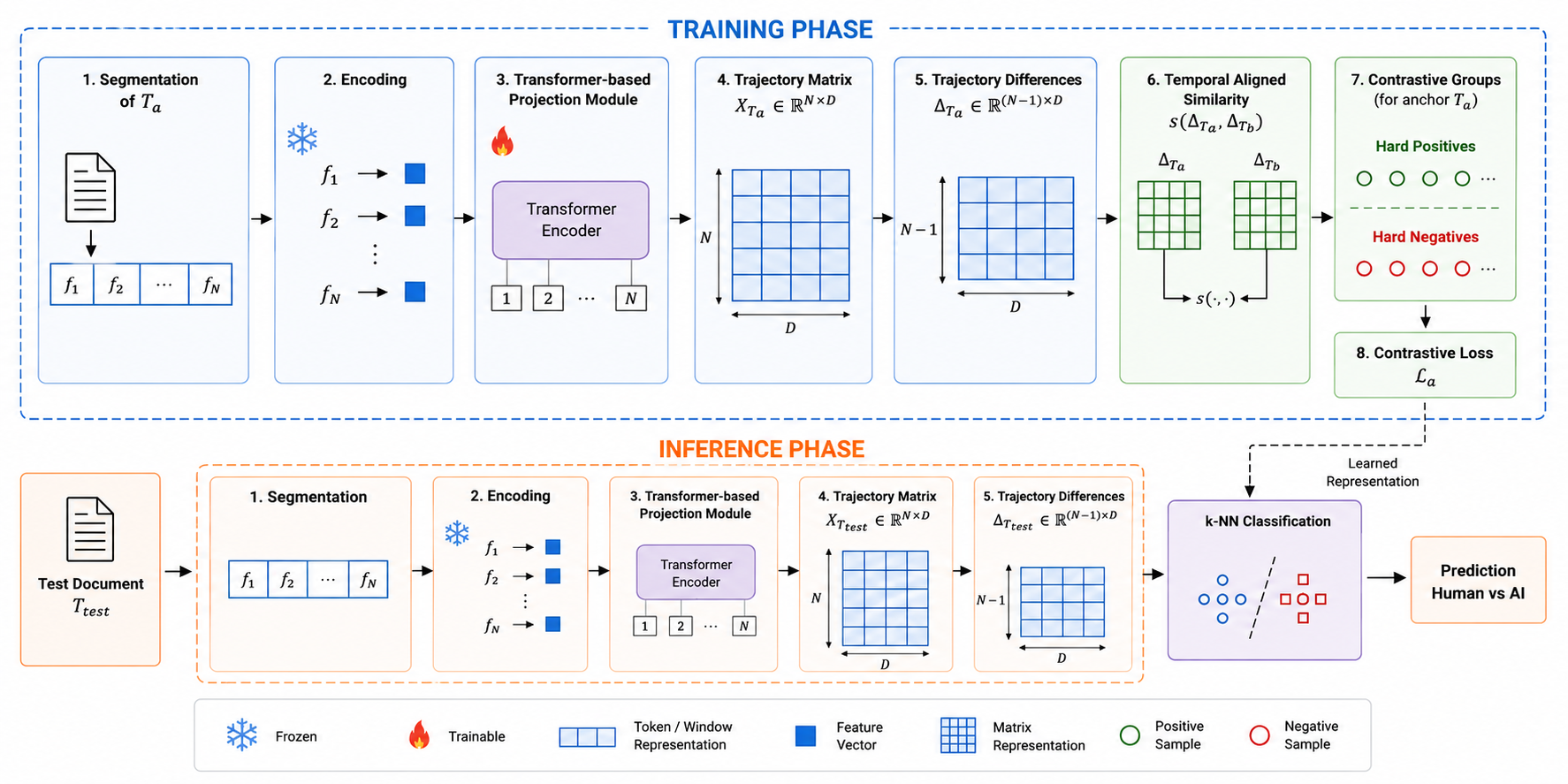}
    \caption{Overview of the \name{} workflow. A document is segmented into overlapping token windows and encoded into a latent trajectory; supervised contrastive learning is applied to the trajectory differences (training), and a k-NN classifier labels documents as human- or AI-generated (inference).}
    \label{fig:framework}
\end{figure}

A graphical representation of the \name{} workflow is provided in Figure \ref{fig:framework}. \name{} behavior consists of three main stages: {\em (i)} trajectory-based text representation, {\em (ii)} supervised contrastive learning over trajectory differences, and {\em (iii)} $k$-nearest neighbors classification in the learned embedding space. During training, a document $T_a$ is segmented into overlapping text windows. These windows are encoded by a frozen encoder $E$ and processed by a trainable transformer-based projection module. This process constructs the trajectory matrix $X_{T_a}$. Then, consecutive trajectory differences $\Delta_{T_a}$ are used to compute temporally aligned similarities and construct hard positive and negative contrastive groups for supervised contrastive learning. During inference, the frozen encoder and projection module generate a document representation that  is classified by a $k$-nearest neighbors classifier. 

\subsection{Steps of the \name{} workflow}
\label{sub:Approach-Underlying-GTCL}

This section outlines the different steps of the \name{} workflow. Each subsection covers one step.

\subsubsection{Text segmentation strategy}
\label{subsub:Text-Segmentation-Strategy}

Let $T = (w_1, w_2, \dots, w_L)$ be a text document composed of an ordered sequence of $L$ words: let $w_i$, $1 \leq i \leq L$, be the $i$-th word of $T$. The goal of this task is to transform $T$ into a structured sequence of local text units while preserving its sequential organization. This allows for the modeling of its internal semantic evolution. To achieve this goal, we use a sliding window mechanism to segment $T$ and extract an ordered sequence of fixed-length text windows. Each window is a contiguous subsequence of words from the original document. We adopted word-based segmentation over token-based ones to maintain architectural agnosticism. Let $M \in \mathbb{N}$ denote the window length (in number of words) and let $Q \in \mathbb{N}$ denote the sliding step between two consecutive windows, with $1 \leq Q \leq M$. Furthermore, let $N \in \mathbb{N}$ be the maximum number of windows associated with each document. This number is fixed a priori to obtain a uniform-size representation. The $i$-th sliding window $f_i$ can be formalized as:

\begin{equation}
    f_i = (w_{r_i}, w_{r_i+1}, \dots, w_{r_i+M-1}),
\end{equation}

\noindent where the starting index $r_i$ is given by:

\begin{equation}
    r_i = (i-1)Q + 1,
\end{equation}

\noindent with $1 \leq i \leq N$.

The parameters $M$ and $Q$ determine the overlap degree between adjacent windows. When $Q < M$, consecutive windows share $M-Q$ words. This allows for the explicit modeling of local continuity and short-range dependencies. When $Q = M$, the windows are contiguous and non-overlapping. Larger values of $Q$ are not meaningful because they would discard intermediate portions of the text. The parameter $N$ ensures that each document is mapped to a structured sequence of a fixed length. For documents that are too short to extract $N$ fully populated windows according to the above definition, boundary conditions are handled in the next representation phase.

Through this segmentation procedure, each document is transformed into an ordered sequence of sliding windows $(f_1, f_2, \dots, f_N)$ that preserves the sequential organization of the original text and provides the basis for constructing a trajectory-based representation in the feature space.

\subsubsection{Construction of the feature vectors}
\label{subsub:Construction-Feature-Vectors}

Once the ordered sequence of text windows $(f_1, f_2, \dots, f_N)$ is obtained, we first pass each window through a pretrained sentence-embedding encoder $E$ to produce a latent representation of it. Then, we feed this representation to a  transformer-based projection module that maps it to the target dimensionality $D$. The resulting vectors are $L2$-normalized to ensure scale-invariant computations. We denote the overall encoding function as:

\begin{equation}
    \phi_E : f_i \mapsto z_i
\end{equation}

\noindent where $z_i = \phi_E(f_i) \in \mathbb{R}^{D}$ is the feature vector associated with the window $f_i$. During training, only the transformer-based projection module is optimized under the contrastive learning objective, while $E$ remains frozen. This design preserves the semantic structure captured by pretraining while enabling the adaptation of the feature space to the downstream discrimination task.

The ordered set of feature vectors $(z_1, z_2, \dots, z_N)$ preserves the sequential structure of the original text and is a structured representation of its internal evolution. Under this formulation, the sequence of feature vectors can be interpreted as a discrete trajectory in the feature space. Each point in this trajectory represents the semantic state of a local text segment, and consecutive points capture the progression of the document.

Since the window length $M$ and the maximum number of windows $N$ are fixed beforehand, for documents that are too short to allow the extraction of $N$ fully populated windows, the last available window is replicated to fill the remaining positions. This ensures that each document is mapped to a sequence of exactly $N$ feature vectors, allowing for direct comparison across texts.

Starting from the vectors of the window features, the document $T$ is represented by the following matrix:

\begin{equation}
    X_T =
    \begin{bmatrix}
        z_1 \\
        z_2 \\
        \vdots \\
        z_N
    \end{bmatrix}
    \in \mathbb{R}^{N \times D}
    \label{eq:trajectory_matrix}
\end{equation}

Each row of $X_T$ encodes the semantic content of a local window. The order of the rows preserves the text's sequential organization. $X_T$ is interpreted as the latent trajectory associated with $T$.

\subsubsection{Determining trajectory dynamics}
\label{subsub:Determining-Trajectory-Dynamics}

Starting from the trajectory matrix $X_T$, we formulate the AIGTD problem as a contrastive learning task over trajectory differences associated with texts of different origins.

Rather than comparing the absolute positions of vectors of window features in the projection space, we focus on the local evolution of trajectories, defined by the sequence of differences between consecutive adjacent windows. Vectors of absolute features primarily reflect the global semantic content, which is heavily influenced by topic and lexical choices. Conversely, a sequence of differences describes how the representation changes over time, isolating the incremental transformations characterizing document progression.

From a geometric perspective, modeling first order differences makes the representation invariant to global translational shifts in the projection space. Any constant offset applied uniformly to all vectors of window features is canceled out by subtraction. This suppresses absolute positional information and emphasizes relative transitions between consecutive semantic states. From a generative perspective, these transitions are linked to the autoregressive generation mechanism of LLMs. Since each token is generated based on the preceding context, adjacent text segments evolve under a continuity constraint imposed by the next token prediction task.

Given $X_T$, we define $\delta_i = z_{i+1} - z_i$, $1 \leq i \leq N-1$. The ordered sequence of differences is represented by the matrix:

\begin{equation}
\Delta_T =
\begin{bmatrix}
\delta_1 \\
\delta_2 \\
\vdots \\
\delta_{N-1}
\end{bmatrix}
\in \mathbb{R}^{(N-1)\times D}
\label{eq:diff_matrix}
\end{equation}

\subsubsection{Computation of temporally aligned similarities}
\label{subsub:Computation-Temporally-Aligned-Similarities}

Let $T_a$ and $T_b$ be two documents, and let $\Delta_{T_a} \in \mathbb{R}^{(N-1)\times D}$ and $\Delta_{T_b} \in \mathbb{R}^{(N-1)\times D}$ be the corresponding sequences of latent trajectory differences. We define the similarity of $\Delta_{T_a}$ and $\Delta_{T_b}$ as a weighted aggregation of the pairwise inner products of their local transitions. Formally speaking:

\begin{equation}
    s\bigl(\Delta_{T_a}, \Delta_{T_b}\bigr) 
    = \sum_{i=1}^{N-1} \sum_{j=1}^{N-1} 
      w_{ij} \, \langle \delta_{a_i},\, \delta_{b_j} \rangle
    \label{eq:causal-sim}
\end{equation}

\noindent where $\delta_{a_i}$ (resp. $\delta_{b_j}$) is the $i$-th (resp., $j$-th) local trajectory difference of $T_a$ (resp., $T_b$), whereas $\langle \delta_{a_i},\, \delta_{b_j} \rangle$ is the inner product of $\delta_{a_i}$ and $\delta_{b_j}$. The weight $w_{ij}$ encodes two structural constraints, i.e., temporal proximity and directional temporal alignment. It is defined as follows:

\begin{equation}
    w_{ij} =
    \begin{cases}
        Z^{-1} e^{-\gamma (i-j)} & \text{if } j \le i \\
        0 & \text{if } j > i
    \end{cases}
    \label{eq:causal-weight}
\end{equation}

\noindent Here, $Z = \sum_{i=1}^{N-1} \sum_{j=1}^{N-1} e^{-\gamma (i-j)} $ is a normalization factor such that $\sum_{i=1}^{N-1} \sum_{j=1}^{N-1} w_{ij} = 1$. $\gamma \in [0,1]$ is a decay parameter that controls the rate at which temporal influence decreases.  Its value determines the degree of temporal locality in the similarity computation. When $\gamma = 0$, all pairs with $j \le i$ are given equal weight. This results in uniform aggregation of local differences and, consequently, in a more global comparison of trajectory dynamics. As $\gamma$ tends to $1$, the exponential term $e^{-\gamma (i-j)}$ induces a progressively stronger decay, concentrating similarity on pairs with $i$ close to $j$ and emphasizing locally aligned differences between trajectories.

The condition $j \le i$ restricts comparisons to temporally compatible local differences and enforces a directional temporal alignment between trajectories. When evaluating the $i$-th local difference $\delta_{a_i}$ of $T_a$, only the differences $\delta_{b_j}$ of $T_b$, occurring at the same or earlier relative positions, are considered. This constraint reflects the sequential nature of text generation. In an autoregressive process, the semantic state at position $i$ depends on preceding states but not on future ones. Since each $\delta_i$ encodes the local evolution between consecutive segments, restricting comparisons to indices $j \le i$ ensures that the similarity measure respects the forward progression intrinsic to the generative process and avoids aligning current differences with information corresponding to later stages of another trajectory.

From a structural perspective, this asymmetric masking prevents early-stage differences in one document from being misaligned with late-stage differences in another document. If present, such misalignment would disregard the cumulative and directional nature of semantic development.

\subsubsection{Performing supervised contrastive learning and classifying documents}
\label{subsub:Performing-Classifying}

Based on the temporally aligned similarity defined in Equation~\ref{eq:causal-sim}, we perform supervised contrastive training following the formulation in~\cite{Khosla*20}. During this phase, the backbone encoder $E$ is fixed, while the transformer-based projection module is optimized under the contrastive objective.

Let $\mathcal{TD}$ denote the set of training documents. For each document $T \in \mathcal{TD}$, we compute its trajectory matrix $X_T$ and the associated sequence of local differences $\Delta_T$, as defined in Equation~\ref{eq:diff_matrix}. 

Rather than constructing mini-batches through random sampling, we first generate a collection of contrastive groups designed to emphasize hard positive and negative examples. For each anchor document $T_a$, we compute the similarity  between its difference sequence $\Delta_{T_a}$ and the difference sequences of all the remaining documents in the training set using the $s(\cdot, \cdot)$ function defined in Equation~\ref{eq:causal-sim}. Let $\mathcal{A}_a$ denote the set of the $\psi$ most similar documents belonging to the opposite class (hard negatives), and let $\mathcal{P}_a$ denote the set of the $\psi-1$ most dissimilar documents belonging to the same class (hard positives). The contrastive group associated with $T_a$ is then defined as follows:

\begin{equation}
\mathcal{G}_a = \{T_a\} \cup \mathcal{P}_a \cup \mathcal{A}_a
\label{eq:Ga}
\end{equation}

We denote the value $|\mathcal{G}_a|$ by $\Psi$. This value is also equal to $2\psi$.

Consequently, each group contains hard negative and hard positive examples, encouraging the model to distinguish between highly similar documents from different classes and reduce the distance between highly dissimilar documents from the same class. After generating all groups, duplicate ones are removed, yielding the final collection of contrastive groups used for training. Given a contrastive group $\mathcal{G}_a$, the supervised contrastive loss for $T_a$ is defined as follows:

\begin{equation}
\mathcal{L}_a =
- \frac{1}{|\mathcal{P}_a|}
\sum_{p \in \mathcal{P}_a}
\log \left(
\frac{
e^{\frac{s(\Delta_{T_a}, \Delta_{T_p})}{\tau}}
}{
\displaystyle\sum_{n \in \mathcal{G}_a \setminus \{T_a\}}
e^{\frac{s(\Delta_{T_a}, \Delta_{T_n})}{\tau}}
} \right)
\label{eq:infonce}
\end{equation}

Here, $\tau > 0$ is a temperature parameter that controls the sharpness of the similarity distribution. The numerator considers all hard positive examples within the group and the denominator includes every other example except the anchor itself. The overall objective is obtained by averaging $\mathcal{L}_a$ over all valid anchors in the group, and then over all contrastive groups.

After contrastive training is complete, the transformer-based projection module $\phi_E$ is frozen. Finally, each text window $f_i$ is projected into the learned representation space $z_i$ through the frozen model $\phi_E$. Rather than training an additional classifier, document classification is performed using a $k$-nearest neighbors ($k$-NN) classifier that operates directly on the learned representations. This approach aligns with the objective of supervised contrastive learning, which is to promote the generation of compact neighborhoods of samples from the  same class in the embedding space while separating samples from different classes. 

\section{Experiments}
\label{sec:Experiments}

This section describes the experiments conducted to evaluate GTCL. Specifically, Subsection \ref{sub:Experimental-Setup} presents the experimental setup. Subsection \ref{subsec:validation_trajectory} assesses the effectiveness of the latent trajectory-based strategy, which represents the core of GTCL. Subsection \ref{sub:quantitative} evaluates the performance of GTCL and compares it with other related systems proposed in the previous literature. Subsection \ref{sub:hyperparameter_analysis} presents an analysis of the hyperparameters characterizing GTCL. Finally, Subsection \ref{sub:ablation_study} illustrates an ablation study related to GTCL. 

\subsection{Experimental Setup}
\label{sub:Experimental-Setup}

In this section, we present our experimental setup; in particular, we describe the models, the datasets, the performance metrics, and the hyperparameter values adopted for our tests. We conducted the experiments using the pre-trained encoder-only transformer Nomic Embed Text v1.5\cite{Nussbaum*25} as the backbone $E$ (see Section \ref{subsub:Construction-Feature-Vectors}).

We performed the evaluation task on three datasets with different characteristics. The first dataset is a subset of RAID \cite{Dugan*24}, an adversarial benchmark designed to evaluate the robustness of detectors against paraphrasing and rewriting strategies. We constructed this subset by randomly selecting human- and AI-generated texts from the original benchmark. In particular, we selected the AI-generated samples from the outputs produced by widely adopted, instruction-tuned, conversational LLMs (i.e., ChatGPT, LLaMA-Chat, and Mistral-Chat) commonly used in real-world applications. The second dataset is a randomly selected subset of NYT-AI \cite{Roy*25}, which comprises human-authored New York Times articles and synthetic counterparts generated by six LLMs (i.e., Gemma-2-9B, Mistral-7B, Qwen-2-72B, LLaMA-8B, Yi-Large, and GPT-4o) covering a diverse range of open-source and proprietary models with varying architectures and scales. The third dataset is Reviews\footnote{\url{https://github.com/christopherburatti/GTCL-AIDetection}}, that we constructed specifically for this study. It consists of human- and AI-generated reviews of scientific papers. We collected the human reviews from the OpenReview\footnote{\url{https://openreview.net/}} peer review process, while we produced the AI-generated reviews by prompting Gemini 1.5 Flash, Gemini 2.0 Flash, and GPT-4o-mini to review the same papers. This dataset is challenging due to the technical and domain-specific nature of the text and to the fact that AI-generated reviews are based on real scientific content, which makes superficial stylistic differences less noticeable. 

To ensure a fair comparison across datasets, we preprocessed all texts by removing special characters and formatting artifacts that LLMs might have introduced during generation. These artifacts include markdown symbols, bullet point indicators, and non-standard punctuation; if not removed, they could act as spurious discriminative features. We divided each dataset in two subsets, namely a training subset (comprising 80\% of the elements) and a testing subset (encompassing 20\% of the elements). Following DeTeCtive \cite{Guo*24}, we then merged the training partitions of all datasets  to train \name{}; afterwards, we evaluated our framework on the corresponding test partition of each dataset separately. Table~\ref{tab:datasets} shows the main characteristics of the datasets considered.

\begin{table}[ht!]
\centering
{\scriptsize
\begin{tabular}{lcccc}
\toprule
{\em Dataset} & {\em Total number} & {\em Number of} & {\em Number of} & {\em Average length} \\
& {\em of documents} & {\em human-generated documents} & {\em AI-generated documents} & {\em (number of words)} \\
\midrule
RAID  & 1,827 & 960 & 867 & 1,570 \\
NYT-AI & 14,000 & 2,000 & 12,000 & 975 \\
Reviews & 2,716 & 1,552 & 1,164 & 2,072 \\
\bottomrule
\end{tabular}
}
\caption{Main features of the datasets used in our experiments.}
\label{tab:datasets}
\end{table}

We measured performance using Accuracy and weighted F1-score. We also reported separate F1-scores for the human-generated and AI-generated classes, to analyze potential asymmetries in discriminatory power. We compared \name{} with several baselines representing the main detection paradigms. In particular, we considered: Desklib\footnote{\url{https://huggingface.co/desklib/ai-text-detector-v1.01}},  Target Mining RoBERTa (TMR)\footnote{\url{https://huggingface.co/Oxidane/tmr-ai-text-detector}}, ModernBERT\cite{DrYiLa25}, Binoculars \cite{Hans*24}, RADAR \cite{HuChHo23}, and DeTeCtive \cite{Guo*24}.

We used the following default values for the GTCL's hyperparameters: {\em (i)} sliding step $Q=8$; {\em (ii)} window length $M=64$; {\em (iii)} number of windows $N=32$; {\em (iv)} decay factor $\gamma=0.2$; {\em (v)} nearest neighbor number $k=9$; {\em (vi)} contrastive group size $\Psi=128$.

The code for GTCL's implementation is publicly available at \url{https://github.com/christopherburatti/GTCL-AIDetection}.

\subsection{Validation of the latent trajectory-based strategy}
\label{subsec:validation_trajectory}

Before evaluating our \name{} framework, we provide statistical evidence that the trajectory differences of human- and AI-generated texts exhibit structurally distinct geometric properties. To this end, we computed the values of several trajectory-level parameters for the sequence of window embeddings associated with each document. We computed the values of these parameters and compared their distributions across the two classes through the Mann--Whitney U test \cite{MaYa16} performed on the three datasets of our interest. We recall that the Mann--Whitney U test is a non-parametric test that assesses whether two distributions differ significantly without making distributional assumptions. A $p$-value less than $0.05$ in this test indicates statistically significant differences between the two classes along the corresponding geometric dimension. 

Let $e_1, e_2, \dots, e_N \in \mathbb{R}^D$ denote the ordered sequence of window embeddings obtained through $E$ and associated with a document. This sequence constitutes a discrete trajectory in embedding space. Let $\tilde{\delta}_i = e_{i+1} - e_i$ denote the difference between the $(i+1$)-th and $i$-th embeddings. We chose a set of geometric metrics over this trajectory. Each metric produces a single scalar value per document; starting from these values we computed the mean and standard deviation of each metric across all documents within each class. The metrics we chose are the following:

\begin{itemize}

\item {\em Trajectory length}: It measures the total distance traveled in the embedding space. It is formalized as:

\begin{equation}
    \Lambda = \sum_{i=1}^{N-1} \|\tilde{\delta}_i\|_2
\end{equation}

where $\|\tilde{\delta}_i\|_2$ denotes the $i$-th step size obtained by computing the corresponding L2 norm. A high value of $\Lambda$ indicates that the text covers a wider region of the latent space, exhibiting larger or more frequent shifts between consecutive segments.

\item {\em Step irregularity}: It measures how much the magnitude of consecutive embedding differences varies within a single document. Let $\rho_i = \|\tilde{\delta}_i\|_2$ denote the $i$-th step size and let $\bar{s} = \frac{1}{N-1}\sum_{i=1}^{N-1} \rho_i$  denote the mean of all step sizes. Step irregularity is formalized as:

\begin{equation}
    \mathcal{I} = \sqrt{\frac{1}{N-1}\sum_{i=1}^{N-1}(\rho_i - \bar{s})^2}
\end{equation}

A high value of ${\cal I}$ indicates that the text alternates between phases of slow and locally coherent progressions and sudden semantic jumps, while a low value reflects a uniform pace of semantic change throughout the document.

\item {\em Curvature}: It measures the tortuosity of the trajectory relative to the straight line displacement between its endpoints. It is formalized as:

\begin{equation}
    \kappa = \frac{\Lambda - \|e_N - e_1\|_2}{\Lambda}
\end{equation}

where $\|e_N - e_1\|_2$ denotes the size of the straight line from $e_1$ to $e_N$. A low value of $\kappa$ indicates a perfectly straight trajectory, while a high value of this parameter corresponds to a highly convoluted trajectory. A high curvature value indicates that the text bends through the latent space rather than progressing linearly from its opening to its closing semantic states. This reflects digressions, perspective shifts, and non-linear semantic development.

\item{\em Directional dispersion}: It measures how much the orientation of consecutive embedding differences fluctuates along the sequence. Let $\hat{v}_i = \tilde{\delta}_i / \|\tilde{\delta}_i\|_2$ be the unit normalization of the $i$-th step vector. The turning angle at position $i$ is defined as:

\begin{equation}
    \theta_i = \arccos\!\left(\hat{v}_i \cdot \hat{v}_{i+1}^\top \right), 
    \quad i = 1, \dots, N-2
\end{equation}

where $\theta_i = 0$ if the $i$-th and $(i+1)$-th differences point in the same direction, while $\theta_i = \pi$ if they point in opposite directions. Directional dispersion is defined as:

\begin{equation}
    \mathcal{D} = \sqrt{\frac{1}{N-2}\sum_{i=1}^{N-2}(\theta_i - \bar{\theta})^2}
\end{equation}

where $\bar{\theta} = \frac{1}{N-2}\sum_{i=1}^{N-2} \theta_i$. A high value of ${\cal D}$  indicates that the orientation of consecutive differences varies widely and irregularly along the sequence, while a low value of this parameter reflects a directionally consistent progression through the latent space.

\end{itemize}

For each dataset, Table~\ref{tab:trajectory_stats} reports the mean and standard deviation of each metric defined above computed separately for human- and AI-generated texts. We also computed the $p$-values of the Mann--Whitney U test and obtained that they were consistently less than 0.05 for all tests conducted on all datasets. The results show the existence of statistically significant differences across all metrics and datasets, which provides empirical support for the hypothesis that human- and AI-generated texts exhibit structurally distinct trajectory differences.

\begin{table}[ht!]
\centering
\scriptsize
\setlength{\tabcolsep}{5pt}
\begin{tabular}{l l cc }
\toprule
& & \textit{Human-generated} & \textit{AI-generated} \\
\cmidrule(lr){3-3} \cmidrule(lr){4-4}
{\em Dataset} & {\em Metric} & {\em Mean value} $\pm$ {\em Standard deviation} & {\em Mean value} $\pm$ {\em Standard deviation} \\
\midrule
\multirow{5}{*}{Reviews}
 & Trajectory length      & $2.648 \pm 0.762$ & $2.189 \pm 0.512$  \\
 & Step irregularity      & $0.054 \pm 0.010$ & $0.043 \pm 0.008$  \\
 & Curvature              & $0.892 \pm 0.061$ & $0.871 \pm 0.033$  \\
 & Directional dispersion & $0.531 \pm 0.173$ & $0.510 \pm 0.117$  \\
\midrule
\multirow{5}{*}{NYT-AI}
 & Trajectory length      & $1.887 \pm 0.537$ & $1.727 \pm 0.190$  \\
 & Step irregularity      & $0.062 \pm 0.018$ & $0.052 \pm 0.005$  \\
 & Curvature              & $0.866 \pm 0.072$ & $0.846 \pm 0.027$  \\
 & Directional dispersion & $0.331 \pm 0.061$ & $0.319 \pm 0.040$  \\
\midrule
\multirow{4}{*}{RAID}
 & Trajectory length      & $2.336 \pm 0.776$ & $2.400 \pm 0.699$  \\
 & Step irregularity      & $0.039 \pm 0.011$ & $0.035 \pm 0.012$  \\
 & Curvature              & $0.888 \pm 0.046$ & $0.900 \pm 0.043$  \\
 & Directional dispersion & $0.487 \pm 0.196$ & $0.447 \pm 0.206$  \\
\bottomrule
\end{tabular}
\caption{Trajectory geometry statistics for human- and AI-generated texts on the Reviews, NYT-AI, and RAID datasets. Each cell reports the mean value $\pm$ standard deviation for documents within each of the two classes of interest.}
\label{tab:trajectory_stats}
\end{table}

As for the Reviews dataset, human-generated trajectories\footnote{Here and in the following, we use the term ``human-generated trajectories'' (resp., ``AI-generated trajectories'') to denote the trajectories of the human-generated (resp., AI-generated) documents.} are longer and more tortuous than AI-generated ones, indicating that human writing explores the semantic space in a less constrained way. The curvature difference confirms that human texts tend to meander through the latent space rather than progress along a directed path. Directional dispersion and step irregularity further suggest that consecutive embedding differences are more variable in human texts, which exhibit phases of stable progression that alternate with more abrupt semantic shifts. 

As for the NYT-AI dataset, which includes a variety of source models and domains, all four metrics demonstrate the same directional pattern observed in the Reviews dataset. Human-generated trajectories are longer, more curved, and more irregular than AI-generated ones. This pattern's consistency across a multi-model corpus suggests that the geometric regularity of AI-generated trajectories is not specific to a single model family but rather reflects a general structural property of autoregressive generation. 

As for the RAID dataset, which is the most challenging adversarial benchmark, directional dispersion and step irregularity preserve the trends observed in the other datasets, confirming their robustness under adversarial conditions. However, trajectory length and curvature show a partial inversion. In fact, AI-generated texts exhibit slightly longer and less tortuous trajectories than human-generated ones. This is consistent with the effect of adversarial attacks, which alter the local geometric structure of text without fully eliminating the discriminative signal. 

Generally speaking, human-generated trajectories are more spread out than AI-generated ones across all datasets, indicating that human writing is geometrically more diverse at the document level. AI-generated trajectories tend to cluster in a smaller region of latent space due to the autoregressive generation mechanism, which imposes local continuity constraints producing more regular and predictable transitions between consecutive semantic states.

These findings provide direct empirical support for the design of \name{}. The geometric differences between human- and AI-generated trajectories, which are observable across all datasets, demonstrate that transitions between consecutive embedding states carry discriminative information about the generative process. In the GTCL framework,  these transitions are precisely what the differences between consecutive vectors of window features capture. Therefore, it is worthwhile to explicitly model the sequential evolution of the trajectory rather than compressing a document into a single static vector.

\subsection{Evaluation of GTCL's performance}
\label{sub:quantitative}

We evaluated the performance results obtained by \name{} and compared them with those of the related approaches mentioned in Section 4.1. Table~\ref{tab:qualitative_results} reports our results. 

\begin{table}[ht!]
\centering
{\scriptsize
\begin{tabular}{llcccc}
\toprule
{\em Dataset} & {\em Approach} & {\em Accuracy} & {\em F1-score} & {\em F1-score (Human)} & {\em F1-score (AI)} \\
\midrule
\multirow{7}{*}{Reviews}
 & \name & \textbf{0.98} & \textbf{0.98} & \textbf{0.98} & \textbf{0.97} \\
 & Binoculars & 0.57 & 0.47 & 0.72 & 0.23 \\
 & Desklib & 0.71 & 0.68 & 0.79 & 0.54 \\
 & DeTeCtive & \underline{0.90} & \underline{0.90} & \underline{0.90} & \underline{0.90} \\
 & ModernBERT & 0.53 & 0.51 & 0.63 & 0.35 \\
 & RADAR & 0.53 & 0.42 & 0.68 & 0.08 \\
 & TMR & 0.47 & 0.34 & 0.13 & 0.62 \\
\midrule
\multirow{7}{*}{NYT-AI}
 & \name & \textbf{0.98} & \textbf{0.98} & \textbf{0.91} & \textbf{0.99} \\
 & Binoculars & 0.43 & 0.48 & 0.33 & 0.51 \\
 & Desklib & 0.92 & 0.91 & 0.67 & \underline{0.96}\\
 & DeTeCtive & \underline{0.94} & \underline{0.94} & \underline{0.78} & \underline{0.96} \\
 & ModernBERT & 0.63 & 0.68 & 0.16 & 0.77 \\
 & RADAR & 0.67 & 0.72 & 0.30 & 0.79 \\
 & TMR & 0.86 & 0.80 & 0.07 & 0.92 \\
\midrule
\multirow{7}{*}{RAID}
 & \name & \textbf{0.83} & \textbf{0.82} & \textbf{0.83} & \textbf{0.82} \\
 & Binoculars & 0.75 & 0.72 & 0.81 & 0.63 \\
 & Desklib & 0.78 & 0.78 & 0.79 & 0.79 \\
 & DeTeCtive & \underline{0.81} & \underline{0.81} & \underline{0.82} & \underline{0.81} \\
 & ModernBERT & 0.74 & 0.74 & 0.77 & 0.71 \\
 & RADAR & 0.65 & 0.62 & 0.71 & 0.50 \\
 & TMR & 0.66 & 0.65 & 0.59 & 0.72 \\
\bottomrule
\end{tabular}
}
\caption{Comparison of the performance of GTCL and related approaches on the three datasets. For each metric and dataset, the highest value is highlighted in bold.}
\label{tab:qualitative_results}
\end{table}

As for the Reviews dataset, \name{} consistently achieves the best performance across all evaluation metrics. Specifically, it outperforms the second-best approach (i.e., DeTeCtive) by 8.89\% in Accuracy, weighted F1-score, and F1-score for human-generated documents, as well as by 7.78\% in F1-score for AI-generated documents. These results confirm that the sliding window representation captures useful information and that contrastive optimization is a key factor to guarantee robust discrimination.

As for the NYT-AI dataset, \name{} achieves the best overall performance. It outperforms the second-best approach (i.e., DeTeCtive) by 4.26\% in Accuracy and weighted F1-score, 16.67\% in F1-score for human-generated documents, and 3.13\% in F1-score for AI-generated documents. The advantage of \name{} is particularly evident in the human-generated class, where competing approaches exhibit significant performance degradation. DeTeCtive achieves the strongest baseline performance, but its F1-score for human-generated documents is substantially lower than the \name{}'s one (0.78 versus 0.91).

As for the RAID dataset, which is the most challenging adversarial benchmark, \name{} consistently achieves the best performance across all evaluation metrics. In particular, it outperforms the second-best approach (i.e., DeTeCtive) by 2.47\% in Accuracy, 1.23\% in weighted F1-score, and 1.22\% in F1-score for human-generated documents, and 1.23\% in F1-score for AI-generated documents. Although the margin of improvement is smaller than on the other two datasets, these results demonstrate the effectiveness of the proposed trajectory-based representation and supervised contrastive learning framework even under more challenging generation settings.

Results from all three datasets show that consecutive differences in latent representations capture the sequential dynamics of autoregressive generation more effectively than static or aggregate signals.

\subsection{Hyperparameter Analysis}
\label{sub:hyperparameter_analysis}

In this experiment, we aimed to analyze how individual hyperparameters impact the GTCL's performance. To this end, we varied each hyperparameter independently, while keeping all the others fixed at their default values. This procedure ensured that any observed performance variations were attributable only to the hyperparameter under examination. We recall that the hyperparameters and their default values are as follows (see Section \ref{sub:Experimental-Setup}): $Q=8$, $M=64$, $N=32$, $\gamma=0.2$, $\Psi=128$, $k=9$.  We conducted the analysis on the three datasets used for the other tests. The results obtained reveal consistent patterns across all three datasets. 

Regarding the sliding step $Q$ (Table~\ref{tab:step}), performance increases monotonically up to the optimal value and then decreases slightly. Small values of $Q$ produce heavily overlapping windows, introducing redundancy in the representation of the trajectory; in fact, consecutive deltas are nearly identical in this case, which decreases the dynamic signal available for contrastive learning. Conversely, too high values of $Q$ reduce the overlap between adjacent windows, which increases the semantic distance between consecutive points on the trajectory, resulting in the loss of relevant local transitions.

\begin{table}[ht!]
\centering
{\scriptsize
\begin{tabular}{llcccc}
\toprule
{\em Dataset} & $Q$ & {\em Accuracy} & {\em F1-score} & {\em F1-score (Human)} & {\em F1-score (AI)} \\
\midrule
\multirow{5}{*}{Reviews}
 & 1 & 0.81 & 0.80 & 0.85 & 0.75 \\
 & 2 & 0.86 & 0.86 & 0.88 & 0.84 \\
 & 4 & 0.91 & 0.91 & 0.93 & 0.89 \\
 & \textbf{8} & \textbf{0.98} & \textbf{0.98} & \textbf{0.98} & \textbf{0.97} \\
 & 16 & 0.96 & 0.96 & 0.97 & 0.95 \\
\midrule
\multirow{5}{*}{NYT-AI}
 & 1 & 0.97 & 0.97 & 0.88 & 0.98 \\
 & 2 & 0.97 & 0.97 & 0.89 & 0.98 \\
 & 4 & 0.97 & 0.97 & 0.88 & 0.98 \\
 & \textbf{8} & \textbf{0.98} & \textbf{0.98} & \textbf{0.91} & \textbf{0.99} \\
 & 16 & \textbf{0.98} & 0.97 & 0.90 & 0.98 \\
\midrule
\multirow{5}{*}{RAID}
 & 1 & 0.66 & 0.66 & 0.69 & 0.63 \\
 & 2 & 0.65 & 0.61 & 0.74 & 0.46 \\
 & 4 & 0.63 & 0.60 & 0.53 & 0.69 \\
 & \textbf{8} & \textbf{0.83} & \textbf{0.82} & \textbf{0.83} & \textbf{0.82} \\
 & 16 & 0.82 & 0.81 & 0.81 & 0.81 \\
\bottomrule
\end{tabular}
}
\caption{Performance of \name{} against changes in the sliding step $Q$}
\label{tab:step}
\end{table}

Performance remains relatively stable as the window length $M$ varies (Table~\ref{tab:window_length}), with a preference for intermediate values. Too short windows tend to capture incomplete semantic units, resulting in noisy embeddings that are less representative of the local content. Conversely, too long windows aggregate excessively heterogeneous portions of text, flattening the semantic differences between consecutive windows and reducing the discriminatory power of GTCL.

\begin{table}[ht!]
\centering
{\scriptsize
\begin{tabular}{llcccc}
\toprule
{\em Dataset} & $M$ & {\em Accuracy} & {\em F1-score} & {\em F1-score (Human)} & {\em F1-score (AI)} \\
\midrule
\multirow{3}{*}{Reviews}
 & 32 & \textbf{0.98} & \textbf{0.98} & 0.97 & \textbf{0.97} \\
 & \textbf{64} & \textbf{0.98} & \textbf{0.98} & \textbf{0.98} & \textbf{0.97} \\ 
 & 128 & 0.96 & 0.95 & 0.96 & 0.95 \\
\midrule
\multirow{3}{*}{NYT-AI}
 & 32 & 0.94 & 0.93 & 0.83 & 0.92 \\
 & \textbf{64} & \textbf{0.98} & \textbf{0.98} & \textbf{0.91} & \textbf{0.99} \\
 & 128 & 0.97 & 0.97 & 0.89 & 0.96 \\
\midrule
\multirow{3}{*}{RAID}
 & 32 & 0.74 & 0.73 & 0.74 & 0.72 \\
 & \textbf{64} & \textbf{0.83} & \textbf{0.82} & \textbf{0.83} & \textbf{0.82} \\
 & 128 & 0.77 & 0.76 & 0.78 & 0.74 \\
\bottomrule
\end{tabular}
}
\caption{Performance of \name{} against changes in the window length $M$}
\label{tab:window_length}
\end{table}

A similar trend is observed for the number of windows $N$ (Table~\ref{tab:num_windows}). Insufficient document coverage results in a trajectory that is too short to capture the geometric regularities associated with autoregressive generation. Conversely, exceeding the optimal value includes redundant or poorly informative portions of text, which dilutes the overall discriminative signal.

\begin{table}[ht!]
\centering
{\scriptsize
\begin{tabular}{llcccc}
\toprule
{\em Dataset} & $N$ & {\em Accuracy} & {\em F1-score} & {\em F1-score (Human)} & {\em F1-score (AI)} \\
\midrule
\multirow{3}{*}{Reviews}
 & 16 & 0.92 & 0.91 & 0.92 & 0.90 \\
 & \textbf{32} & \textbf{0.98} & \textbf{0.98} & \textbf{0.98} & \textbf{0.97} \\
 & 64 & \textbf{0.98} & 0.97 & 0.97 & \textbf{0.97} \\
\midrule
\multirow{3}{*}{NYT-AI}
 & 16 & 0.94 & 0.94 & 0.89 & 0.95 \\
 & \textbf{32} & \textbf{0.98} & \textbf{0.98} & \textbf{0.91} & \textbf{0.99} \\
 & 64 & 0.96 & 0.96 & 0.84 & 0.98 \\
\midrule
\multirow{3}{*}{RAID}
 & 16 & 0.65 & 0.64 & 0.57 & 0.69 \\
 & \textbf{32} & \textbf{0.83} & \textbf{0.82} & \textbf{0.83} & \textbf{0.82} \\
 & 64 & 0.82 & 0.81 & \textbf{0.83} & 0.77 \\
\bottomrule
\end{tabular}
}
\caption{Performance of \name{} against changes in the number of windows $N$}
\label{tab:num_windows}
\end{table}

Finally, the hyperparameter $\gamma$ (Table~\ref{tab:gamma}) controls the degree of temporal locality in the similarity measure between trajectories. Performance degrades for too low or too high values of $\gamma$. Indeed too low values of this hyperparameter are equivalent to making a global comparison of trajectories, which disregards the local structure of transitions. Conversely, too high values of $\gamma$ restrict the comparison excessively to the same time, ignoring the medium-range dependencies characterizing the semantic progression of the text. The optimal value of $\gamma$ lies within an intermediate range and is robust across all the three analyzed datasets.

\begin{table}[ht!]
\centering
{\scriptsize
\begin{tabular}{llcccc}
\toprule
{\em Dataset} & $\gamma$ & {\em Accuracy} & {\em F1-score} & {\em F1-score (Human)} & {\em F1-score (AI)} \\
\midrule
\multirow{6}{*}{Reviews}
 & 0 & 0.91 & 0.91 & 0.92 & 0.89 \\
 & 0.1 & 0.94 & 0.94 & 0.95 & 0.92 \\
 & \textbf{0.2} & \textbf{0.98} & \textbf{0.98} & \textbf{0.98} & \textbf{0.97} \\
 & 0.3 & 0.97 & 0.97 & \textbf{0.98} & 0.96 \\
 & 0.5 & 0.95 & 0.94 & 0.94 & 0.94 \\
 & 1 & 0.89 & 0.89 & 0.90 & 0.89 \\
\midrule
\multirow{6}{*}{NYT-AI}
 & 0 & 0.93 & 0.92 & 0.90 & 0.93 \\
 & 0.1 & 0.95 & 0.95 & 0.90 & 0.96 \\
 & \textbf{0.2} & \textbf{0.98} & \textbf{0.98} & \textbf{0.91} & \textbf{0.99} \\
 & 0.3 & 0.96 & 0.96 & \textbf{0.91} & 0.97 \\
 & 0.5 & 0.94 & 0.93 & 0.89 & 0.93 \\
 & 1 & 0.91 & 0.91 & 0.85 & 0.91 \\
\midrule
\multirow{6}{*}{RAID}
 & 0 & 0.75 & 0.75 & 0.75 & 0.75 \\
 & 0.1 & 0.80 & 0.80 & 0.81 & 0.79 \\
 & \textbf{0.2} & \textbf{0.83} & \textbf{0.82} & \textbf{0.83} & \textbf{0.82} \\
 & 0.3 & 0.77 & 0.77 & 0.77 & 0.77 \\
 & 0.5 & 0.75 & 0.73 & 0.76 & 0.70 \\
 & 1 & 0.70 & 0.70 & 0.69 & 0.71 \\
\bottomrule
\end{tabular}
}
\caption{Performance of \name{} against changes in the decay factor $\gamma$}
\label{tab:gamma}
\end{table}

Table~\ref{tab:ablation_group_size} analyzes the impact of contrastive group size $\Psi$ (see Section \ref{subsub:Performing-Classifying}) on \name{} performance. As the table shows, very small groups ($\Psi=16$ or $32$) generally lead to slightly lower performance. This is likely because they provide a less informative set of hard positive and hard negative examples during contrastive learning. Increasing the group size improves results. The best overall performance is obtained with $\Psi=128$ across all three datasets. A further increase to $\Psi=256$ does not provide additional benefits and sometimes slightly degrades performance. This suggests that excessively large groups may introduce less informative or noisier samples.

\begin{table}[ht!]
\centering
{\scriptsize
\begin{tabular}{llcccc}
\toprule
{\em Dataset} & $\Psi$ & {\em Accuracy} & {\em F1-score} & {\em F1-score (Human)} & {\em F1-score (AI)} \\
\midrule
\multirow{5}{*}{Reviews}
 & 16 & 0.93 & 0.93 & 0.94 & 0.93 \\
 & 32 & 0.94 & 0.94 & 0.94 & 0.93 \\
 & 64 & 0.96 & 0.96 & 0.96 & 0.95 \\
 & \textbf{128} & \textbf{0.98} & \textbf{0.98} & \textbf{0.98} & \textbf{0.97} \\
 & \textbf{256} & \textbf{0.98} & \textbf{0.98} & \textbf{0.98} & \textbf{0.97} \\
\midrule
\multirow{5}{*}{NYT-AI}
 & 16 & 0.97 & 0.97 & 0.90 & 0.98 \\
 & 32 & 0.97 & 0.97 & 0.90 & 0.98 \\
 & 64 & 0.97 & 0.97 & \textbf{0.91} & 0.98 \\
 & \textbf{128} & \textbf{0.98} & \textbf{0.98} & \textbf{0.91} & \textbf{0.99} \\
 & 256 & 0.97 & 0.97 & 0.89 & 0.98 \\
\midrule
\multirow{5}{*}{RAID}
 & 16 & 0.78 & 0.78 & 0.78 & 0.78 \\
 & 32 & 0.75 & 0.75 & 0.76 & 0.75 \\
 & 64 & 0.80 & 0.80 & 0.80 & 0.81 \\
 & \textbf{128} & \textbf{0.83} & \textbf{0.82} & \textbf{0.83} & \textbf{0.82} \\
 & 256 & 0.79 & 0.79 & 0.81 & 0.76 \\
\bottomrule
\end{tabular}
\caption{Performance of \name{} against changes in the group size $\Psi$}
\label{tab:ablation_group_size}}
\end{table}

Table~\ref{tab:ablation_knn} reports the performance of \name{} for different values of $k$ in the $k$-nearest neighbors classifier (see Section \ref{subsub:Performing-Classifying}). As can be seen from the table, \name{} is robust to the choice of $k$, exhibiting only marginal performance variation across all three datasets. This behavior suggests that the learned representation creates stable local neighborhoods, enabling the $k$-nearest neighbors classifier to produce consistent predictions across a wide range of neighborhood sizes.

\begin{table}[ht!]
\centering
\scriptsize
\begin{tabular}{llcccc}
\toprule
{\em Dataset} & $k$ & {\em Accuracy} & {\em F1-score} & {\em F1-score (Human)} & {\em F1-score (AI)} \\
\midrule
\multirow{5}{*}{Reviews}
 & 5 & 0.97 & 0.96 & 0.96 & 0.95 \\
 & 7 & 0.97 & 0.97 & 0.97 & 0.96 \\
 & \textbf{9} & \textbf{0.98} & \textbf{0.98} & \textbf{0.98} & \textbf{0.97} \\
 & 11 & 0.97 & 0.97 & 0.97 & 0.96 \\
 & 13 & 0.96 & 0.96 & 0.96 & 0.96 \\
\midrule
\multirow{5}{*}{NYT-AI}
 & 5 & 0.96 & 0.96 & 0.89 & 0.98 \\
 & 7 & 0.96 & 0.96 & 0.90 & 0.98 \\
 & \textbf{9} & \textbf{0.98} & \textbf{0.98} & \textbf{0.91} & \textbf{0.99} \\
 & 11 & 0.97 & 0.97 & 0.90 & 0.98 \\
 & 13 & 0.96 & 0.96 & 0.89 & 0.98 \\
\midrule
\multirow{5}{*}{RAID}
 & 5 & 0.80 & 0.80 & 0.81 & 0.80 \\
 & 7 & 0.81 & 0.81 & 0.82 & 0.81 \\
 & \textbf{9} & \textbf{0.83} & \textbf{0.82} & \textbf{0.83} & \textbf{0.82} \\
 & 11 & 0.81 & 0.81 & 0.82 & 0.81 \\
 & 13 & 0.80 & 0.80 & 0.81 & 0.80 \\
\bottomrule
\end{tabular}
\caption{Performance of \name{} against changes in the number of neighbors}
\label{tab:ablation_knn}
\end{table}

\subsection{Ablation Study}
\label{sub:ablation_study}

Our ablation study aimed to evaluate the robustness of \name{} in relation to the selected backbone encoder. To this end, we tested \name{} using Jina Embeddings V3 \cite{Sturua*24} as an alternative to Nomic Embed Text v1.5. Jina Embeddings V3 is a 570-million-parameter model. It is substantially larger and more powerful than the 137-million-parameter Nomic Embed Text v1.5 model used in previous experiments. Table~\ref{tab:ablation} shows the results obtained with both backbones across the three datasets. 

\begin{table}[ht!]
\centering
{\scriptsize
\begin{tabular}{llcccc}
\toprule
{\em Dataset} & {\em Backbone} & {\em Accuracy} & {\em F1-score} & {\em F1-score (Human)} & {\em F1-score (AI)} \\
\midrule
\multirow{2}{*}{Reviews}
 & Nomic Embed Text v1.5 & 0.98 & 0.98 & 0.98 & 0.97 \\
 & Jina Embeddings V3    & 0.97 & 0.96 & 0.96 & 0.95 \\
\midrule
\multirow{2}{*}{NYT-AI}
 & Nomic Embed Text v1.5 & 0.98 & 0.98 & 0.91 & 0.99 \\
 & Jina Embeddings V3    & 0.98 & 0.97 & 0.89 & 0.98 \\
\midrule
\multirow{2}{*}{RAID}
 & Nomic Embed Text v1.5 & 0.83 & 0.82 & 0.83 & 0.82 \\
 & Jina Embeddings V3    & 0.84 & 0.82 & 0.83 & 0.82 \\
\bottomrule
\end{tabular}
}
\caption{Performance of \name{} using Nomic Embed Text v1.5 and Jina Embeddings V3 as backbone encoders}
\label{tab:ablation}
\end{table}

From the analysis of this table, we can observe that the performance is consistent across the two configurations, confirming that the discriminative signal captured by \name{} is independent of the specific backbone encoder. These results suggest that the contrastive objective based on trajectories learns geometric regularities that are robust to the choice of the underlying embedding model. They also suggest that GTCL can be effectively instantiated with backbones of different scales and architectures. This implies that Nomic Embed Text v1.5 is sufficient as a backbone encoder for GTCL. As a consequence, there is no need for larger, and therefore more expensive, encoders.

\section{Discussion}
\label{sec:Discussion}

The experimental results reported in the previous section demonstrate that explicitly modeling the sequential dynamics of latent representations yields a signal that can discriminate between AI-generated and human-generated text. \name{}'s consistent advantage across all datasets suggests that trajectory-based contrastive learning captures the structural properties of the generative process. Notably, \name{} maintains balanced detection across both classes, a property several existing approaches fail to preserve. 

\name{}'s innovation lies in modeling text as a sequence of evolving representations. While virtually all existing methods treat documents as static objects that can be summarized into a single representation, \name{} treats documents as trajectories in an embedding space. In this space, the discriminative signal emerges from the geometry of semantic transitions rather than from aggregate content. This is not merely a technical variation, but rather a conceptual reframing of the detection problem. Instead of observing what the text looks like, \name{} learns how the text evolves. The empirical evidence presented in Section~\ref{subsec:validation_trajectory} supports this hypothesis by showing that the geometric properties of trajectory differences differ significantly between human- and AI-generated texts across datasets. 

\name{} offers several practical advantages over existing AIGTD paradigms. First, it operates exclusively on text, requiring no access to logits or the generative model's internal states. This makes \name{} applicable to any text, regardless of its origin. Second, \name{} produces well-calibrated predictions for both human- and AI-generated texts. 

This property is especially valuable in real-world deployment scenarios, where minimizing the costs associated with both false positives and false negatives is essential.

\section{Conclusion}
\label{sec:Conclusion}

In this paper, we proposed \name{}, an AI-Generated Text Detection (AIGTD) framework whose novelty lies in modeling the evolution of representations within a document. In fact, rather than treating a document as a static object to be summarized into a single representation, \name{} treats it as a discrete trajectory in latent space. In this space, the discriminative signal emerges from the geometry of semantic transitions between consecutive local segments. This conceptual reframing, from static representation to sequential dynamics, is the paper's central contribution. It is based on the observation that the autoregressive generation mechanism induces structural regularities in the evolution of latent representations that are not observed in human-generated documents.

First, GTCL segments each document into ordered, overlapping windows. Then, it encodes each window using a frozen backbone encoder, followed by a transformer-based projection module. Finally, it applies supervised contrastive learning directly to the sequence of consecutive embedding differences. A temporally aligned similarity measure, designed to consider the forward progression of the generative process, structures the latent space according to the geometric dynamics of the trajectory, rather than global content similarity. 

The evaluation of GTCL on three different benchmarks demonstrates that the novel perspective underlying it yields robust and discriminative signals across models and domains. Furthermore, the geometric properties of trajectory differences constitute a structurally grounded and empirically validated information source for detection.

A promising direction for future work is improving the interpretability of GTCL. Since the discriminative signal is distributed along the trajectory, the development of attribution methods identifying the windows or transitions that contribute most to the classification decision could reveal which parts of the text are most indicative of artificial generation. This could lead to new approaches for the explainable detection of AI-generated text. Another area for future research concerns the extension of GTCL to texts produced by diffusion-based language models. Unlike autoregressive models, diffusion-based ones generate text through iterative denoising over the entire sequence, rather than by sequential left-to-right prediction. This different mechanism may induce distinct geometric properties in trajectory differences, which would require the adoption of a powered version of GTCL to account for the absence of a direction in the generative process.


\section*{Declaration of competing interest}
The authors declare that they have no known competing financial interests or personal relationships that could have appeared to influence the work reported in this paper.

\section*{Data availability}
The datasets and code used for our study are publicly available at \url{https://github.com/christopherburatti/GTCL-AIDetection}.

{\footnotesize

\bibliographystyle{plain}
\bibliography{bibliografia,temp}

}

\end{document}